\documentclass[sigconf,anonymous=false]{acmart}
\usepackage{graphicx}
\usepackage{subfiles}

\usepackage{todonotes}
\usepackage{booktabs}
\PassOptionsToPackage{hyphens}{url}\usepackage{hyperref}

\usepackage{dsfont}
\usepackage{amsmath}

\usepackage{caption}
\usepackage{subcaption}

\usepackage{mfirstuc}
\usepackage{enumitem}%
\setlist[itemize]{align=parleft,left=0pt..1em}

\usepackage{xcolor}

\newcommand{\deletioncandidate}[1]{#1}

\newcommand{\sourceelement}{report segment}

\newcommand{\targetelement}{target element}

\newcommand{\clrg}{ARGUE}
\newcommand{\reportrequest}{report request}
\newcommand{\capreportrequest}{Report Request}

\AtBeginDocument{%
  \providecommand\BibTeX{{%
    \normalfont B\kern-0.5em{\scshape i\kern-0.25em b}\kern-0.8em\TeX}}}

\copyrightyear{2024}
\acmYear{2024}
\setcopyright{rightsretained}
\acmConference[SIGIR '24]{Proceedings of the 47th International ACM SIGIR Conference on Research and Development in Information Retrieval}{July 14--18, 2024}{Washington, DC, USA}
\acmBooktitle{Proceedings of the 47th International ACM SIGIR Conference on Research and Development in Information Retrieval (SIGIR '24), July 14--18, 2024, Washington, DC, USA}
\acmDOI{10.1145/3626772.3657846}
\acmISBN{979-8-4007-0431-4/24/07}

\makeatletter
\gdef\@copyrightpermission{
 \begin{minipage}{0.3\columnwidth}
 \href{https://creativecommons.org/licenses/by/4.0/}{\includegraphics[width=0.90\textwidth]{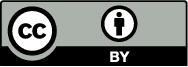}}
 \end{minipage}\hfill
 \begin{minipage}{0.7\columnwidth}
 \href{https://creativecommons.org/licenses/by/4.0/}{This work is licensed under a Creative Commons
Attribution International 4.0 License.}
 \end{minipage}
 \vspace{5pt}
}
\makeatother

\begin{document}

\title{On the Evaluation of Machine-Generated Reports}

\renewcommand{\shortauthors}{James Mayfield et al.}

\begin{abstract}
Large Language Models (LLMs) have enabled new ways to satisfy information needs.
Although great strides have been made in applying them to settings like document ranking and short-form text generation,
they still struggle to compose complete, accurate, and verifiable long-form reports.
Reports with these qualities are necessary to satisfy the complex, nuanced, or multi-faceted information needs of users.
In this perspective paper, we draw together opinions from industry and academia,
and from a variety of related research areas, %
to present our vision for automatic report generation, and---critically---a flexible framework by which such reports can be evaluated.
In contrast with other summarization tasks,
automatic report generation starts with a detailed description of an information need,
stating the necessary background, requirements, and scope of the report.
Further, the generated reports should be complete, accurate, and verifiable.
These qualities, which are desirable---if not required---in many analytic report-writing settings,
require rethinking how to build and evaluate systems that exhibit these qualities.
To foster new efforts in building these systems,
we present an evaluation framework that draws on ideas 
found in various evaluations. %
To test completeness and accuracy, the framework uses nuggets of information, expressed as questions and answers, that need to be part of any high-quality generated report.
Additionally, evaluation of citations that map claims made in the report to their source documents ensures verifiability.

\end{abstract}

\begin{CCSXML}
<ccs2012>
   <concept>
       <concept_id>10002951.10003317</concept_id>
       <concept_desc>Information systems~Information retrieval</concept_desc>
       <concept_significance>500</concept_significance>
       </concept>
 </ccs2012>
\end{CCSXML}

\ccsdesc[500]{Information systems~Information retrieval}

\keywords{Evaluation, Report Generation, Text Analysis, Factual Citation}

\author{James Mayfield}
\affiliation{%
  \institution{Johns Hopkins University}
  \city{Baltimore, MD}
  \country{USA}
}
\email{mayfield@jhu.edu}

\author{Eugene Yang}
\affiliation{%
  \institution{Johns Hopkins University}
  \city{Baltimore, MD}
  \country{USA}
}
\email{eugene.yang@jhu.edu}

\author{Dawn Lawrie}
\affiliation{%
  \institution{Johns Hopkins University}
  \city{Baltimore, MD}
  \country{USA}
}
\email{lawrie@jhu.edu}

\author{Sean MacAvaney}
\affiliation{%
  \institution{University of Glasgow}
  \city{Glasgow}
  \country{United Kingdom}
}
\email{sean.macavaney@glasgow.ac.uk}

\author{Paul McNamee}
\affiliation{%
  \institution{Johns Hopkins University}
  \city{Baltimore, MD}
  \country{USA}
}
\email{mcnamee@jhu.edu}

\author{Douglas W. Oard}
\affiliation{%
  \institution{University of Maryland}
  \city{College Park, MD}
  \country{USA}
}
\email{oard@umd.edu}

\author{Luca Soldaini}
\affiliation{%
  \institution{Allen Institute for AI}
  \city{Seattle, WA}
  \country{USA}
}
\email{luca@soldaini.net}

\author{Ian Soboroff}
\affiliation{%
  \institution{NIST}
  \city{Gaithersburg, MD}
  \country{USA}
}
\email{ian.soboroff@nist.gov}

\author{Orion Weller}
\affiliation{%
  \institution{Johns Hopkins University}
  \city{Baltimore, MD}
  \country{USA}
}
\email{oweller2@jhu.edu}

\author{Efsun Kayi}
\affiliation{%
  \institution{Johns Hopkins University}
  \city{Baltimore, MD}
  \country{USA}
}
\email{ekayi1@jhu.edu}

\author{Kate Sanders}
\affiliation{%
  \institution{Johns Hopkins University}
  \city{Baltimore, MD}
  \country{USA}
}
\email{ksande25@jhu.edu}

\author{Marc Mason}
\affiliation{%
  \institution{Johns Hopkins University}
  \city{Baltimore, MD}
  \country{USA}
}
\email{mmason8@jhu.edu}

\author{Noah Hibbler}
\affiliation{%
  \institution{University of Maryland}
  \city{College Park, MD}
  \country{USA}
}
\email{ndhibbl@terpmail.umd.edu}

\renewcommand{\shortauthors}{J. Mayfield et al.}

\maketitle

\section{Introduction}\label{sec:introduction}

The emergence of generative Large Language Models (LLMs) has brought with it
the ability to automatically generate all kinds of text.
With it, a host of problems---old and new---have (re)emerged that affect these generated texts.
The fields of Information Retrieval (IR) and Natural Language Processing (NLP) both have important roles in building new methods to improve text generation and in designing approaches to evaluate the quality of these methods.

LLMs can enable new ways for people to satisfy various information needs.
Simple information needs (e.g., factoids) can be answered with relatively short generated responses pointing to a single source.
However, when information needs are complex, nuanced, or multifaceted,
a suitable response must also be more complex.
They need to draw together numerous facts gathered from potentially multiple sources
to completely and faithfully respond to the information need.
We refer to this longer-form answer generation as a ``report'' on a user-specified topic.

More formally, we define a report as a text that attempts to satisfy an explicitly stated information need
by finding documents in a corpus (potentially a mixture of text, images, tables, etc.) that contain relevant information,
expressing that information in the text, 
and providing appropriate citations from the report to the supporting documents.
We envision a high-quality report as the \textit{ideal} response to a user with a complex task in mind,
since such a report would succinctly, coherently, and verifiably cover all the information in a corpus
pertinent to their information need.
Note that this definition makes the framework better suited to reports that inform an analyst
than to reports that generate novel analyses.

Report writing can be viewed as a natural downstream task of Retrieval Augmented Generation (RAG), where faithfulness has been a focus of study~\cite{ragas, saad2023ares, llamaindex}.
In this view, an LLM generates the report
using the report request as part of the prompt
and searches the document collection for relevant information that can be added to the prompt
to ensure the report's accuracy.
Report generation can also be thought of as summarization.
From the summarization viewpoint, a report is an attributed task-based informative abstractive multi-document summary
(see Section~\ref{sec:summarization} for a more detailed explanation of these categories).
Such a report might also include portions that are not summaries at all,
but are, for example, introductory material or comparisons of the summarized information.

We posit that all of these viewpoints are valid,
and each informs evaluation for report generation.
This work describes an abstract framework for evaluating automated report generation, \clrg\
(Automated Report Generation Under Evaluation),
that is built on top of lessons learned from prior evaluation approaches in information retrieval,
summarization and text generation. 
It will be used by the TREC track NeuCLIR in its report generation task.\footnote{\url{https://neuclir.github.io/}}
The \clrg\ framework builds a foundation for a broader research agenda in evaluating automatically generated long-form text beyond reports.

Some of \clrg's most important features are:
\begin{itemize}[noitemsep,topsep=0pt]

    \item We use the concept of information \textbf{nuggets} out of the summarization literature
    to capture the content a report should contain.
    We express each nugget as a question together with a list of acceptable answers to that question.

    \item \textbf{Citations} are a key report component.
    A citation is a pointer from a source element in the report
    (typically a sentence)
    to a target element in a document
    (typically the entire document).

    \item
    We propose that precision and recall serve as the basis for most content-based measures.
    \clrg\ supports precision measures over the sentences of the report,
    and recall measures over the information nuggets.

\end{itemize}

\section{Requirements} \label{sec:requirments}

This section defines requirements of a report evaluation system.

We first define the various actors (and one non-actor) in \clrg:

\subparagraph{\textit{Report Requester:}} The person requesting the report. This is the person whose purpose the report should satisfy.

\subparagraph{\textit{Report Audience:}} The person who will be reading the report.
    This is often
    the same as the report requester.

\subparagraph{\textit{Report Writer:}} The automated system that takes a \reportrequest\ and a document collection as inputs
    and produces the report.

\subparagraph{\textit{\capreportrequest:}} A detailed specification of the report to be written.
    The \reportrequest\ can include:
    \begin{itemize}[noitemsep,topsep=0pt]
        \item \textit{User story:} explains the report requester's background, situation, and report-writing philosophy,
        as well as a description of the audience for the report.
        \item \textit{Problem statement:} indicates the content that the report is required to contain. 
        \item \textit{Background}: describes what is already known about the topic that need not appear in the report.
        \item \textit{Constraints:} specifies restrictions such as the length of the report or a temporal window for sources.
    \end{itemize}

\subparagraph{\textit{Assessor:}} Any person making judgments in producing evaluation materials
    or scoring submitted runs.
    Assessors include those selecting report topics,
    writing report requests, 
    identifying nuggets,
    binding nuggets to documents in the collection,
    and making other judgments necessary to assign scores to reports.

The evaluation we advocate has several key attributes.
First, it must ensure that the report is responsive to the report request.
It must ensure the report's key information presented is attested in the document collection,
that the report properly cites those documents,
and that the information they contain is faithfully captured by the report.
It must score a report using evaluation data created by a person.
While scoring may be automated,
requiring the ground truth data to be human-generated
helps to prevent circularity between report generation and report evaluation,
thereby reducing the bias the evaluation might have toward e.g., a particular generative model.
Finally, the evaluation must have the intention of reusability. 
Producing a reusable evaluation is challenging
because of the level of interpretation required to make the required judgments.
Reusability is thus often at odds with the other goals of an evaluation.
The information retrieval community has thought through many of the issues underlying reusability,
and we present \clrg\ to try to take advantage of that experience.

While it is nearly impossible to accurately claim that any evaluation component is novel,
there are points of emphasis in our proposed evaluation style
that we think make it stand out from other extant text generation evaluations.
First is the amount and detail of the background information provided in the report request.
While other evaluations have provided additional information describing inclusion criteria,
in practice systems have often focused only on brief specifications.
For example, a narrative giving detailed information about what should and should not be considered relevant,
long a part of TREC topics,
has rarely been exploited.
The arrival of large language models that can easily incorporate such materials
makes now an opportune time to focus on including ancillary documentation in a report request,
not just for this style of evaluation,
but for any text generation evaluation.
\deletioncandidate{While we advocate that these ancillary details be made explicit in the evaluation,
we acknowledge that in real report-writing applications implicit knowledge might be more practical
and adequate for the task.}

Second, until recently hallucination in text generation system output was not a major focus,
primarily because generative systems were not good enough to create convincing hallucinated text.
With the rise of large generative LLMs
hallucination has become a common part of text generation system output;
the evaluation must account for this as well.

Borrowing from an IR evaluation perspective,
we promote the view of nuggets as opinion, not fact.
In report evaluation, nuggets play the role that relevant documents play in IR.
Were document relevance treated as fact rather than opinion,
it would be virtually impossible to come to agreement on which documents were relevant to a given topic;
inter-annotator agreement would be too low.
Treating relevance as opinion avoids this problem.
In exchange, relevance as opinion adds constraints to the evaluation,
primarily that the author of the topic should be the relevance assessor.
If relevance is not decided until after system submissions,
that means that assessor continuity is important;
assessors should be selected such that they can create topics at one time,
and assess relevance at a later time, possibly months later.
We advocate accepting this tradeoff for nuggets in report generation evaluation.
For nuggets, the implication is that items reasonably seen by a report writer as nuggets
might not be identified in advance
by the assessor.
A given evaluation might address this issue through a pyramid approach~\cite{pyramid} to identify nugget importance
if multiple reference reports are available.
Or an evaluation might determine that nugget subjectivity will not change the preference order
of meaningfully different systems and ignore it.
In either case, we recommend that report sentences bearing and accurately reflecting a citation
should not be penalized during scoring,
precisely because they might be valid nuggets in someone's eyes.
Constraints such as maximum document length can discourage intentional overgeneration
of sentences that have a small chance of matching assessor nuggets.

To meet these requirements,
four broad questions should be asked about each report being evaluated:
\begin{enumerate}[noitemsep,topsep=0pt]
    \item[Q1]
    Does the report include the information contained in the document collection that the report
    requires?

    \item[Q2]
    Does it accurately 
    express all such information?

    \item[Q3]
    Does it contain appropriate citations to the collection?

    \item[Q4]
    Has the information been fitted together into a useful form?

\end{enumerate}

\noindent Q4 is a crucial part of any text generation evaluation.
It covers such attributes as fluency~\cite{mauve}, coherence~\cite{coherence-lapata,coherence-lin}, consistency~\cite{true}, and rhetorical structure~\cite{dias-rhetoricalstructure, mello-rhetoricalstructure}.
In light of this importance, it has a long history and has been studied in depth elsewhere.
Thus, while we leave a place for this in the overall evaluation in \clrg,
we leave it to others to address it in light of the changing NLP landscape.

\section{Background}\label{sec:background}

Here we review related work on report writing and evaluation.

\subsection{Report Writing} \label{sec:reportwriting} %

Report writing involves text generation, for which prior work on summarization and RAG provides useful perspectives.

\subsubsection{Summarization} \label{sec:summarization} 

In its most general form, a summary is a document whose substantive content is based entirely on the content of other \textit{target document(s),}
and that is more concise than simply presenting the other document(s) in their original form would have been~\cite{mani2001automatic}.
Summaries have been defined along several axes:
\begin{itemize}[noitemsep,topsep=0pt]
    \item Single-document or Multi-document~\cite{lin2002single}:
    Is the summary built from one document (single-document),
    or many (multi-document)?
    \item Extractive or Abstractive~\cite{chopra2016abstractive}:
    Does the summary primarily draw language from the summarized documents (extractive),
    or does it generate new language (abstractive)?
    \item Indicative or Informative~\cite{klavans2001domain}:
    Does the summary help the reader to decide whether to read the summarized document(s) (indicative),
    or does it include enough content to make it unnecessary to read those document(s) (informative)?
    \item Generic or Task-Based~\cite{vanderwende2007beyond}:
    Is the summary constructed with no particular task in mind (generic),
    or is there a specific task that the summary is designed to support (task-based)?
    \item Attributed or Unattributed~\cite{rashkin2023measuring}:
    Does the summary include citations to the summarized documents (attributed),
    or does it lack citations (unattributed)?
    \item Original or Update~\cite{mccreadie2014incremental,park2020continual}:
    Should the summary include all information (original),
    or only information that the reader does not already know (update)?
    \item Closed or Open Domain~\cite{giorgi2023open,zhou2023odsum}:
    Are the documents to summarize supplied (closed domain),
    or must the system perform a search to identify the appropriate documents (open domain)?
\end{itemize}
The reports in which we are interested are attributed task-based informative abstractive open-domain multi-document summaries
that may call for either original or update summaries.

\subsubsection{Retrieval-Augmented Generation} \label{sec:RAG}

Following preliminary research on furnishing transformer architectures with external knowledge sources,
\citet{lewis2021retrievalaugmented} introduce RAG models
as a way to improving language model performance on knowledge-intensive tasks,
using an encoded Wikipedia collection as a non-parametric memory system.
RAG models have since been used to improve dialogue systems~\cite{komeili2021internet, shuster2021retrieval},
machine translation~\cite{cai2021neural, bulte2019neural},
and text-style transfer~\cite{li2018delete}
among other applications~\cite{li2022survey}. 

Various approaches have been proposed to incorporate RAG models into summarization~\cite{peng2019text, an2021retrievalsum}
and other document generation tasks.
One use of retrieval has been to find an example summary,
sometimes with retrieved summary reranking~\cite{cao2018retrieve},
to serve as a template for the summary of another document.
Retrieval can also be used to improve language model factuality. 
By curating large, high quality collections,
generation can be grounded in supporting documents~\citep{barham2023megawika}.
This mechanism has been shown to be particularly beneficial for rarer entities and concepts~\cite{Mallen2022WhenNT}. 
Finally, RAG enables LLMs to access information that was not available at pre-training time,
such as proprietary or copyrighted information~\cite{Min2023SILOLM}.

\deletioncandidate{
Vision-language modeling~\cite{liu2023visual, awadalla2023openflamingo, achiam2023gpt}
enables %
multimodal retrieval-augmented generation systems
that benefit from rich non-textual data 
~\cite{hu2023reveal,moro2023fashion}.
Different modalities facilitate the completion of different tasks, 
including 
image understanding~\cite{chen2022murag, zhu2022visualize},
open-domain VQA~\cite{lin2022retrieval, hu2023reveal},
translation~\cite{fang2022neural},
and multimodal generation~\cite{yasunaga2022retrieval}.
}

\subsection{Evaluation}  \label{sec:eval}

As report generation includes elements of several prior tasks,
including document retrieval, summarization, question answering, and retrieval-augmented generation,
we briefly review salient work on those tasks that we see as related to \clrg.

\subsubsection{Information Retrieval} \label{sec:eval:IR}
Evaluation of ad hoc retrieval is typically based on assessor-produced relevance judgments of documents
that are selected by pooling system responses in a shared task,
or sometimes based on active learning~\cite{roegiest2015trec,grossman2016trec}.
Obtaining both good precision and good recall is important in real-world systems,
so commonly used metrics combine both components
(e.g., mean average precision, nDCG \cite{ndcg2002}).
Statistical significance testing can be performed,
for example with Student's $t$-test \cite{smucker07}.

In a report-writing scenario, recall is important
to allow assessment of how comprehensively the report responds to the report request.
Precision is also important for automated report generation;
reports are a type of multi-document synthesis,
and incorporating content from non-pertinent documents can adversely affect the utility of the report.

To create evaluation datasets for report writing,
care must be taken to develop report requests that match information available in the document collection.
If requests are too broadly scoped, or if too much salient information is present in the collection, it will be difficult
(i.e., prohibitively expensive in human labor)
to determine the full set of correct nuggets present in the collection.

\subsubsection{Summarization} \label{sec:eval:summarizaton}

Evaluating automatic summarization can require significant manual effort.
In 2001, NIST initiated the Document Understanding Conference (DUC)
to develop evaluation methods for summarization.
DUC continued until 2007 and then became the summarization track of the Text Analysis Conference (TAC) through 2014.
The DUC/TAC summarization evaluations were notable for having people write summaries manually,
and using those ``model'' summaries (or ``reference texts'') as the jumping-off point for metric development.

The DUC evaluation procedure measured coverage (that is, recall)
through a pairwise comparison between two summaries: the model summary and a ``peer'' summary
(which could be a generated summary or another model).
The model was divided into {\em Elementary Discourse Units} (EDUs),
essentially clauses~\cite{soricut2003sentence,li2020composing} 
while the peer was split on sentence boundaries.
An assessor would match each EDU with the sentences in the peer that contained that information,
and indicate how much of the meaning of the EDU was expressed in the corresponding matched peer units.
Unmarked sentences in the peer were then marked for relevance.
\citet{harman-DUC-variation} found that model summaries from different authors were markedly different,
and that assessors also did not agree on model unit coverage ratings.

Work also began around DUC 2003 on automatic metrics,
specifically comparing the model summary to the peer using word n-gram statistics.
\citet{lin-bleu} looked at the BLEU measure developed for machine translation,
and found that recall on word unigrams correlated better with the DUC assessments than full BLEU scoring,
which incorporates longer n-grams.
Following that, they developed ROUGE~\cite{lin-rouge},
a recall-oriented metric similar to BLEU.
ROUGE has a number of variants depending on how tokens are parsed,
how n-grams are selected and assembled,
and how scores are aggregated across summaries to obtain a system score.
A study by \citet{graham-2015-evaluating} explored a large grid of ROUGE parameters
in comparison with BLEU using data from DUC-2004,
and found that BLEU and ROUGE-2 (2-grams, stemmed, stopwords removed, computing an average of precision scores)
had the highest correlation with human assessment.
ROUGE has been used to evaluate summarization~\cite{lin-rouge},
Long-Form Question Answering (LFQA)~\cite{xu2023critical, krishna2021hurdles} and RAG~\cite{lewis2021retrievalaugmented}.
ROUGE has well-documented problems as an evaluation metric in e.g., summarization~\cite{graham-2015-evaluating} or LFQA~\cite{krishna2021hurdles}.
From our perspective, its main problems as an evaluation metric for report generation are
its requirement for reference reports
(making it expensive),
its poor robustness to hallucination
(making it inaccurate),
and that it does not handle citations
(making it incomplete).

In 2004, \citet{pyramid} proposed the ``Pyramid Method'' for evaluation.
Since comparing generated summaries against a model is subject to the inherent variation in model summaries,
they propose to abstract the model summaries into {\em Summary Content Units} (SCUs).
SCUs are clauses that appear
(with more or less the same meaning)
in multiple model summaries.
They are weighted by the number of model summaries that express them.
Figure~\ref{scu-example} shows an example of two SCUs from parts of four model summaries.

\begin{figure*}
\def\arraystretch{1.5}
\footnotesize
\centering
\begin{tabular}{|p{10cm}|p{4cm}|p{2.6cm}|}
\hline
\begin{minipage}[t]{\linewidth}
A1 In 1998 \underline{two Libyans indicted in 1991} for the Lockerbie bombing were still in Libya. \\
B1 \underline{Two Libyans were indicted in 1991} for blowing up a
Pan Am jumbo jet over Lockerbie, Scotland in 1988. \\
C1 \underline{Two Libyans, accused} by the United States and
Britain of bombing a New York bound Pan Am jet over
Lockerbie, Scotland in 1988, killing 270 people, for 10
years were harbored by Libya who claimed the suspects
could not get a fair trail in America or Britain. \\
D2 \underline{Two Libyan suspects were indicted in 1991.} 
\end{minipage}

     &  

\begin{minipage}[t]{\linewidth}
SCU1 (w=4): \\
two Libyans were officially accused of the Lockerbie bombing \\
A1 [two Libyans]1 [indicted]1 \\
B1 [Two Libyans were indicted]1 \\
C1 [Two Libyans,]1 [accused]1 \\
D2 [Two Libyan suspects were indicted]1
\end{minipage}
     
     & 

\begin{minipage}[t]{\linewidth}
SCU2 (w=3): \\
the indictment of the two Lockerbie suspects was in 1991 \\
A1 [in 1991]2 \\
B1 [in 1991]2 \\
D2 [in 1991.]2
\end{minipage} \\

\hline
\end{tabular}

\vspace{-1em}
\Description{}
\caption{\label{scu-example} A pair of example Summary Content Units.
Four semantically similar sentences from four different model summaries
are grouped into two SCUs highlighting the key facts from those sentences.
From \citet{pyramid}.}
\end{figure*}

In informal usage, SCUs have been referred to as ``nuggets.''
Rather than being a clause, a nugget might be a description of a concept along with how it was expressed in the models.\footnote{See \url{https://tac.nist.gov/publications/2010/presentations/TAC2010_Summ_Overview.pdf} for an example of SCUs as nuggets.}
Subsequent research on the pyramid method has focused on automatic creation and alignment of SCUs.
For example, \citet{gao-etal-2019-automated} performs a dependency parse of the model summary,
then represents individual clauses using vector embeddings.
Nugget fuzziness can be addressed by using hoppers~\cite{hoppers, mitamura2016overview}
to bin together differing descriptions that refer to the same item.

The main difficulties in using nuggets for report evaluation are that
they treat hallucinations (contradictions and misinformation)
exactly the same as content that has no matching nugget,
and that they do not support citations.
We have incorporated nugget-based evaluation into \clrg,
tying nuggets to reports not directly,
but rather through cited documents.

\subsubsection{Question Answering} \label{sec:eval:QA}

\deletioncandidate{
Factoid Question Answering (QA) evaluation typically consists of using accuracy or $F_1$
against a gold standard answer (or answer set)~\cite{voorhees2001trec,dang2007overview,rajpurkar2016squad}.
This type of evaluation has many advantages,
as it can be easily automated and is simple to annotate.
Long-form QA~\cite{fan2019eli5,pang2021quality}
is evaluated similarly to summarization, %
typically with 
automated metrics like ROUGE,
model-based metrics like BERTScore~\cite{zhang2019bertscore} or BLEURT~\cite{sellam2020bleurt},
or human evaluation~\cite{krishna2021hurdles,xu2023critical}.
}

\subsubsection{Retrieval-Augmented Generation} \label{sec:eval:RAG}%

Early retrieval augmented generation systems have been evaluated using task-specific metrics on end-to-end tasks. 
For example, in the context of question answering,
exact match and $F_1$ metrics have been used~\cite{Guu2020REALMRL,lewis2021retrievalaugmented}.
For summarization, ROUGE and BERTScore on reference summaries are common~\cite{giorgi2023open}. 
These approaches have two limitations:
they only measure ability to complete end tasks,
and thus cannot assess intermediate stages or evaluate generation across multiple dimensions; and
they are not well-suited to capture failures that can be introduced by current generative models~\cite{Goyal2022NewsSA}.

More recently, techniques have proposed to more holistically evaluate RAG systems. 
\citet{Gienapp2023EvaluatingGA}~introduce a theoretical framework for evaluating ad hoc generative retrieval. 
\citet{Chen2023BenchmarkingLL}~focus on robustness of RAG systems against various perturbations.
\citet{Thakur2023NoMIRACLKW}~benchmark hallucinations and the ability of RAG systems to identify relevant information for 18 languages. 
Others have introduced benchmarks to measure the ability of RAG systems to provide citations~\cite{Bohnet2022AttributedQA,Gao2023EnablingLL,Liu2023EvaluatingVI,Yue2023AutomaticEO}.
While not specifically designed for RAG applications,
metrics designed to evaluate factuality (e.g.,  \textit{FactScore}~\cite{Min2023FActScoreFA})
or faithful manipulation of long inputs (e.g., \textit{BooookScore}~\cite{Chang2023BooookScoreAS})
can complement application-specific evaluation frameworks.

Most approaches to automated evaluation aim to estimate the effectiveness of RAG systems across desirable dimensions
(e.g., faithfulness, answer relevance, and context relevance). 
Techniques include prompting LLMs to evaluate generated summaries~\cite{ragas},
and fine-tuning lightweight models on synthetic data~\cite{saad2023ares}.
Downstream applications, such as question answering,
can also be used to evaluate the effectiveness of RAG systems~\cite{exam}.

\section{Proposed Framework}\label{sec:framework}

This section describes our conceptual evaluation framework for automated report generation. 
We name this abstract framework \clrg\  (Automated Report Generation Under Evaluation) for convenience. 
We model the information need as a \textit{report request},
which is analogous to the \textit{topics} in TREC-style ad hoc retrieval evaluation.
The \textit{report writer} is required to respond with a verifiable \textit{report},
with \textit{citations} to its information sources.
As in retrieval system evaluation,
we restrict the system to citing documents in a \textit{pre-defined document collection}
instead of arbitrary information on the web.
The framework is thus limited in the range of writing types it can evaluate.
In particular, it does not currently support evaluation of reported information
that is not explicitly supported by statements in the document collection.
This restriction allows experiments that compare systems across research studies and over time.

\subsection{Framework Overview}

In \clrg, creating a report generation benchmark has three phases.
The first phase creates evaluation data.
We believe that systems should be evaluated over human-curated data
so that they are ranked on effectiveness rather than alignment to machine output.

System input comprises a document collection
and report requests that describe information needs. %
The second phase distributes these inputs to participants. 
Generated reports are expected to be responsive to the information needs statements. 
A valid report will cite source documents that contain the reported information. 
Citations are a key attribute of this framework.
Other report stylistic requirements might include, for example,
a length limit 
to encourage systems to express information succinctly.
If the document collection is in a language different from the report request, or is multilingual,
the report may be required to be written in the language of the report request.
We envision that the input data will be distributed as part of an evaluation campaign,
but this is not required. 
Assuming an evaluation campaign, generated reports will be received and evaluated by assessors;
however, to support reusability,
key components will be replaced by automated mechanisms
to allow future systems to be scored using the same evaluation data.

The third phase scores reports.
Since the goal of this framework is to evaluate systems,
each system will need to generate multiple reports based on the various report requests. 
Report scores will be aggregated to assign system scores. 
Required information in reports will be expressed by assessors in the form of nugget questions and answers.
Answers will be attested in the collection and tied to particular documents that attest those answers,
thereby tying the nuggets to supporting documents.
During scoring, report citations will be used to determine which nuggets are described in the report. 
Thus there will be a notion of \textit{recall} over nuggets, which is a new feature in RAG evaluation.
Citations will also be used to ensure that non-required information that is included in the report
(facts that are not part of the necessary nuggets)
is attested in the collection.
A \textit{precision} score over \textit{report segments} measures how well the report adheres to information found in the collection.
This allows hallucination to be addressed,
whether it be false information or true information that is unattested.
While traditional recall and precision are set measures,
they can be modified to account for some nuggets having greater weight than others
or to allow report segments to bear multiple citations.

\subsection{Evaluation Inputs and Outputs} \label{sec:components}
\subsubsection{Evaluation Inputs} \label{sec:input}

The first system input is the collection of items that will be used as source material for the retrieval task.
While these items could be documents written in one or more languages,
it is also possible for the items to be images, videos, audio files, or some combination.
For the reminder of this paper, we will refer to the items as \textit{documents.}
Because of the importance of having citeable units, 
the document collection will be divided into \textit{\targetelement s},
which are typically documents,
but can be smaller units of text such as passages, paragraphs, or sentences,
depending on the needs of the evaluation.
In this paper we will assume that an entire document has been selected as the \targetelement.
Segmentation into \targetelement s should be done once and distributed with the collection
to ensure that all systems are evaluated on an even footing.
The document collection should include documents that contain sufficient information relevant to the desired report.
Following most information retrieval-based evaluations,
documents are assumed to be truthful;
verifying the truthfulness of document contents is orthogonal to and beyond the scope of the framework.
Instead, the framework focuses on citation,
requiring that all reported information cites supporting documents from the evaluation document collection.
Information that cites a document incorrectly or that is missing a required citation is appropriately penalized.

The second system input is a set of assessor-developed information needs
referred to as \textit{report requests.}
A report will be generated for each report request.
Report requests are more extensive and subtler than information needs for previous IR or summarization tasks.
See Section~\ref{sec:requirments} for the full report request description.

Creation of report requests is a complex process that tries to satisfy multiple, sometimes conflicting goals.
It bears many similarities to topic creation for a TREC-style IR evaluation~\cite{trec}.
In topic identification for \clrg,
the topic creator must be familiar both with information retrieval,
    and with any special requirements of the document collection.
    For example, a bilingual document collection would require that the topic creator be at least bilingual.
    A document collection on medical topics would require topic creators
    who were well-versed in the medical domain.

In addition, an IR evaluation typically tries to control the number of documents that are relevant to the topic being developed, in part because doing so can improve reusability.
    An \clrg\ evaluation must control not only the number of documents that contain relevant information,
    but also the number of nuggets and the number of \targetelement s that align to each nugget.
    Having too many items in any of these categories leads to high assessment costs;
    having too few leads to higher score variance and lower ability to distinguish systems.
    That said, assessors need not capture all information that might satisfy the information need.
    It is up to the assessor
    to determine what, in their opinion, is the essential information. %

\subsubsection{Evaluation Output} \label{sec:output}

The report will be generated by an automated report writer.
Reports produced by the report writer should satisfy the constraints listed in Section~\ref{sec:requirments}.
For the purposes of this framework,
we make a convenience assumption that the report requester and the report audience are the same.
As an example, the assessor could have the role of analyst,
with the purpose of the report being to support the process of drawing analytic conclusions.

The generated report will be segmented into \textit{report segments}, either manually or automatically.
For convenience, we will assume in this work that a report segment is a sentence,
but it could be some other well-defined portion of report text.
Finer-grained segments may enable more nuanced distinctions.
Given that precision scores operate over report segments,
and given that automated sentence segmentation is imperfect,
we believe that it is important that the report writer control the segmentation.
Thus, each report must be segmented into sentences by the report writer prior to evaluation.
The evaluation should include guidelines on sentence segmentation.
The report must also include appropriate citations,
pointers from 
source elements (sentences) to \targetelement s (documents).
Each report sentence will bear zero or more citations, as described below.

\subsection{Citations} \label{sec:citations}

Each substantive sentence of a submitted report must cite the document \targetelement(s) from which it was derived.
Which sentences are substantive may vary according to the goals of the evaluation.
A citation then is a pointer from one report segment to one target element.
A given report segment may bear more than one citation,
and a given target element may be cited more than once.
By traversing such citations the evaluation system can map sentences in the report to documents and then to nuggets.
Note that the report writer must know nothing about the nuggets that will be used to evaluate the report;
they are known exclusively to the assessor.
The assessor may choose to require just one citation per sentence,
or, if completeness is to be measured,
all valid and salient citations.

The validity of a citation has three components.
    First, the report segment must be supported by the \targetelement.
    That is, reading the \targetelement\ should verify the sentence's accuracy.
    In a manual evaluation, the assessor decides whether a given sentence is supported by the \targetelement.
    In an automated evaluation, support of a report segment for a \targetelement\ could be measured in several ways.
    The simplest is a semantic match,
    testing whether the semantics of the two texts match.
    A number of such automated metrics are available,
    such as Sentence-BERT~\cite{reimers2019sentence}.
    A more accurate but harder measurement would be whether the \targetelement\ entails the report sentence.
    Entailment has been a component of evaluation sets such as GLUE~\cite{glue} and SUPERGLUE~\cite{superglue},
    and good solutions to the problem have been identified~\cite{rtesurvey}.

    Second, at the same time, the sentence bearing the citation should be responsive to the report request.
    This means that the cited \targetelement\ is linked to a nugget,
    and that the \sourceelement\ provides
    an answer to one of that nugget's questions (see below for nugget questions).
    Thus the acceptability of a nugget answer depends on which document the report cites.
    Again, the assessor will determine whether the \sourceelement\ answers a nugget question.
    One way to automate assessment of responsiveness might be to use
    an automated QA system to find answers to a nugget question,
    then use a semantic matching system
    to determine whether the \sourceelement\ matches one of those answers.

    Third, some evaluations will
    also assess
    whether a talented author in the field of the report would include that citation if they had written the report.
    An evaluation that simply wants all substantive sentences to bear a citation will omit this component;
    a more nuanced evaluation of reports in their final form could include it. %
    In either case, judgments will need to be made
    on which sentences require a citation.
    Cases where no citation is required include 
    introductory sentences,
    background sentences that reflect the problem statement,
    and sentences that summarize other cited sentences.
    If we are interested only in nugget recall,
    we can safely ignore whether sentences ought to have citations.
    But if we are interested in precision,
    we would not like to penalize a report for containing such non-citing sentences
    (except perhaps when measuring the quality of the report as a whole).
    To handle non-citing sentences,
    it must be determined whether the sentence should have a citation.
    If a citation is not needed, the report can be scored as if %
    the identified sentences were not present in the report.

\subsection{Nuggets} \label{sec:nuggets}

The proposed evaluation is centered on \textit{nuggets.}
A nugget is a piece of information that should appear in the report
and that could be expressed in a variety of ways in the document collection. 

\subsubsection{Nugget Definition}

A \textit{nugget} in this framework is a combination of a question and one or more answers to that question
that address some aspect of the report request
and that are expressed in at least one \targetelement\ in the collection.
Nuggets must be expressed at an appropriate level of granularity for the desired report.
If the report answers such a question using appropriate citations into the document collection,
we deem it to have succeeded in identifying that nugget;
evaluation metrics (described in Section~\ref{sec:metrics} below)
can then use statistics over the correctly answered, incorrectly answered, and unanswered nugget questions
to produce a score for a given report.
Answers to nugget questions should express the information that a reasonable person would expect in a report
written in response to the \reportrequest.

The concept of nuggets arose from summarization evaluation~\cite{pyramid}.
New in this framework is the expression of nuggets as questions with allowable answers.
We are interested in evaluation data that can be used to automatically evaluate systems,
much like relevance assessments can be used to evaluate an IR system
even decades after
their creation.
We believe this formulation will be helpful in automating report generation evaluation.

Nuggets need not capture everything any report responding to the \reportrequest\ might legitimately include.
Given that reports by necessity will be shorter than the source documents,
the assessor will determine the required information and express that as nuggets, 
reinforcing the idea that nuggets are opinions instead of facts. 
\deletioncandidate{The set of answers to a nugget question are drawn from all the answers supported by the document collection.
Questions and answers will be in the request language
even if, for example, the source information comes from an image or is in a different language. }

\subsubsection{Nugget Identification}
\label{sec:nuggetid}

Nuggets are identified by the assessor.
Nuggets must be both relevant to the \reportrequest\ and attested in the document collection.
In practice, the assessor could either
look through retrieved documents to identify important aspects of the topic from the \targetelement s,
or identify nuggets a report on the topic ought to include,
then search the document collection to see which are attested.
A combination of both methods could be used.
To ensure reproducibility and enable evaluating recall, it is desirable to identify most (or all) nuggets that should be included. 

In addition to identifying the set of nuggets for a \reportrequest ,
the assessor must also identify each \targetelement\ in the document collection that supports an answer to each nugget.
To do so, the assessor must have both a way to identify \targetelement s that contain nugget-supporting information,
and a way to bind \targetelement s to nugget answers.
The former problem is similar to that faced by many IR collection developers
of ensuring that all or most relevant items have been discovered.
Full collection annotation is not practical for large collections.
Three main techniques for identifying relevant documents
are interactive search, pooling~\cite{pooling,pooling-reliability,pooldepth},
and active learning~\cite{pool-and-AL,AL-to-build-collections,hical}.
Interactive search is simply having a person use any desired tool to identify relevant documents.
In pooling, the assessor judges only documents found in an aggregate of several systems' top results.
Either assessors must have access to systems that together are likely to find most of the relevant documents,
or this step must wait until task participants have submitted their runs.
It is usually desirable to augment the pools manually using interactive search.
In active learning, a classifier identifies relevant documents.
Each time the assessor judges a document,
the classifier is retrained to take the new judgment into account.
Any or all of these techniques might be used to restrict the number of documents
that must be examined during nugget identification.

The second task, assigning \targetelement s to nuggets, is more challenging.
We highlight three challenges here.
First is within-nugget variation.
For example, one nugget answer might be a superset of another,
such as ``June'' versus ``26 June.''
If the more general answer is acceptable,
the more specific answer must be included in the answer set
to distinguish it from an incorrect answer such as ``12 June.'' 
The summarization community introduced \textit{hoppers}~\cite{hoppers}
to capture commonality across descriptions that differ in some details.
For example, two descriptions of a particular natural disaster might indicate different numbers of casualties;
perhaps the descriptions were written at different times
or based on two different information sources.
Whether hopper-style conflation is used for a given evaluation depends on the desired report type.
An overall report on the natural disaster might use hoppers;
a report on how different news services covered the disaster might need to distinguish differing descriptions.
As with decisions on nugget creation,
if hoppers are used,
the choice of hoppers is left to the assessor.

\begin{figure*}[t]
    \centering
    \includegraphics[width=\linewidth]{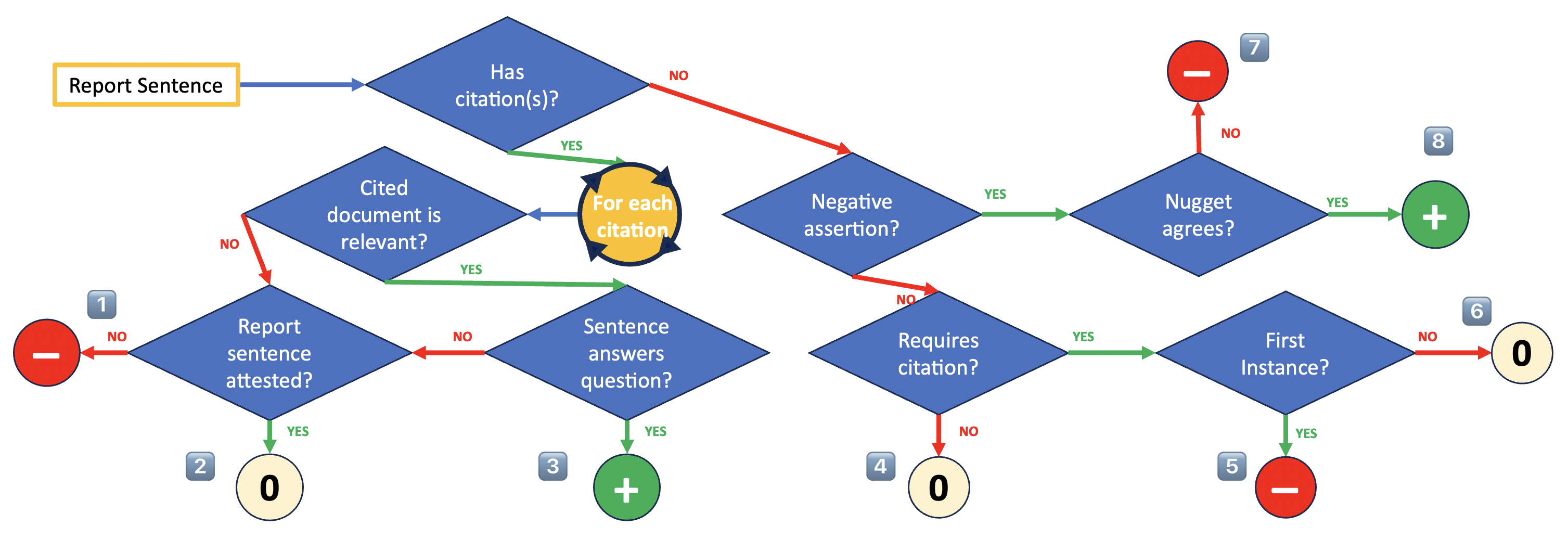}
    \vspace{-2em}
    \caption{Report sentence scoring.
    Answers to eight yes/no questions dictate an outcome for each input sentence.
    + indicates that the sentence is rewarded,
    - that it is penalized,
    and 0 that it does not affect the overall report score.} 
    \label{fig:scoring}
\end{figure*}
\begin{figure*}
\footnotesize

\def\arraystretch{1.5}
\begin{tabular}{|p{7cm}|p{10cm}|}
\hline
\textbf{Report Request}:  
I am a Hollywood reporter writing an article about the highest grossing films Avengers: Endgame and Avatar. My article needs to include when each of these films was considered the highest grossing films and any manipulations undertaken to bring moviegoers back to the box office with the specific goal of increasing the money made on the film. &
\textbf{Gold Standard Report}:
Avatar originally became the highest grossing film in 2010 [D1].
Avengers: Endgame replaced Avatar as the highest grossing film in 2019 
[D1, D2, D3, D8, D10, D12, D13].
It overtook Avatar by adding an additional six minutes of footage to the film to draw viewers back to the movie theater [D4]. Two years later Avatar was re-released in mainland China [D1, D2, D5, D6, D7, D8, D9, D10, D11].
It earned a sufficient amount of money to retake the title of highest-grossing film in 2021 [D5, D11, D6, D7, D2, D8, D9, D1]. 
\end{tabular}

\begin{tabular}{|p{8.5cm}p{8.5cm}|}
\hline
\textbf{Nuggets} as Questions and Answers: 
\begin{enumerate}[leftmargin=*]
    \item When did Avatar first become the highest grossing film?
    \begin{itemize}
        \item 2010 [D1]
    \end{itemize}
    
    \item  When did Avengers: Endgame become the highest grossing film?
    \begin{itemize}
        \item  2019 [D1,D2, D3, D8, D10, D12, D13]
        \item July 2019 [D3, D12, D13]
        \item July 20, 2019 [D3]
        \item July 21, 2019 [D13]$^\dagger$ %
    \end{itemize}
    
    \item What did studio executives do to the Avengers: Endgame film to become the highest grossing film?
    \begin{itemize}
        \item Added six minutes of additional footage [D4]
        \item Added footage [D4]
        \item Added 6 minutes [D4]
        \item Additional footage at the end of the film [D14]
    \end{itemize}
\end{enumerate}\vspace*{-\baselineskip}
& 
\hfill
\begin{enumerate}[leftmargin=*]
    \setcounter{enumi}{3}
    
    \item When did Avatar retake the title of highest grossing film?
    \begin{itemize}
        \item 2021 [D1, D2, D6,D7,D9,D11]
        \item March 2021 [D1, D6 ,D7, D9, D11]
        \item March 13, 2021 [D1, D6, D9]
        \item Two years after the Avengers: Endgame became the highest grossing film [D2]
    \end{itemize}

    \item What event led to Avatar becoming the highest grossing film?
    \begin{itemize}
        \item Re-release in Mainland China [D1, D2, D5, D6, D7, D8, D9, D10]
        \item Re-release in China [D1, D2, D5, D6, D7, D8, D9, D10]
        \item Release in Mainland China for a second time [D1, D2, D5, D6, D7, D8, D9, D10]
        \item Returned to theaters in China [D11]
    \end{itemize}
\end{enumerate} 
\vspace*{1.5\baselineskip}
$^\dagger$In Taiwan Time
\vspace*{-\baselineskip} \\
\hline
\end{tabular}

\vspace{-1em}    
    \caption{Example evaluation material for a report request. }
    \label{fig:example-eval-info}
\end{figure*}

A second challenge is a single \sourceelement\ or \targetelement\ expressing information about more than one nugget.
This is handled through multiple citations borne by a single report sentence,
and/or multiple mappings between \targetelement s and nuggets.
This complicates the bookkeeping needed to give appropriate credit to each nugget,
but poses no theoretical problems.

A third challenge is a single nugget requiring multiple report sentences or \targetelement s to be fully captured.
This challenge arises because
nugget question/answer pairs lend themselves well to simple facts expressed in the report,
but are less well suited to identifying complex information.
Nonetheless we believe that the general framework will be extensible to complex nuggets
whose expression is distributed across several report sentences or \targetelement s
by allowing complex questions answered by Boolean combinations of \targetelement s,
and by exploiting recent research in question answering~\cite{QA-decomposition,complexQA}.

\subsubsection{Practical considerations} 

The following considerations are not requirements of the framework,
but instead practical tips we have gleaned  working to instantiate this and similar evaluation frameworks.
First, we believe that an assessor must be 
familiar both with IR concepts and any special requirements of collection and evaluation topic area
(such as the aforementioned
bilingual or medical settings).
Second, it may be advantageous for an assessor to produce a gold standard report
to help assemble the information that should be in a satisfactory report.
Nugget questions can then be composed from that report.
Creating a gold standard report also enables a ROUGE evaluation for comparison. %
Third,
IR evaluations usually limit the number of relevant documents to simplify and reduce the cost of evaluation.
Report evaluation would also like to control the number of nuggets and document mappings
to ensure the evaluation can distinguish good and bad systems;
however, this can eliminate from consideration practical use cases
that would otherwise be in scope for the task.
This tradeoff has traditionally been considered worthwhile, but it should be remembered that it is a tradeoff.
Fourth, LLMs can call on memorized knowledge not found in the document collection.
Often the LLM training collection is unknown.
If the LLM
has not seen the evaluation corpus, it will need to rely on hallucination,
which will negatively affect evaluation data quality.
Finally, while finding all potential nuggets is unnecessary
since nugget worthiness is an assessor's opinion,
finding all answers to nugget questions is important for collection reusability,
especially as nuggets can only be answered using documents known to have the answer.
If nuggets are generated prior to submissions,
it might be worth pooling submissions to identify more nugget answers.

\subsection{Metrics} \label{sec:metrics}

Many metrics can be used to assess automatically generated reports.
Two common IR measures are recall and precision;
we focus on these here because they are well-known, easy to calculate, 
and highlight most of the important scoring issues we face in generated report evaluation.
\begin{figure*}
\centering
\footnotesize

\begin{tabular}{|p{8.3cm}p{0.1cm}p{8.3cm}|}
\hline
\vspace*{-0.5\baselineskip}
\begin{enumerate}[leftmargin=*]
    \item Avengers: Endgame and Avatar are two of the highest-grossing films in history. \textsc{Outcome: \#4} 
    \item Avengers: Endgame surpassed Avatar as the highest-grossing film globally, with a box office revenue of \$2.787 billion. [D12] \textsc{Outcome: \#2}
    \item This achievement was officially announced in July 2019. [D13] \textsc{Outcome: \#3, Nugget: 2}
    \item To bring moviegoers back to the box office and increase the revenue of Avengers: Endgame, Marvel Studios implemented several strategies. \textsc{Outcome: \#4}
    \item One of these strategies was re-releasing the film with additional content. \textsc{Outcome: \#6}
    \item Marvel Studios President Kevin Feige confirmed plans for a re-release, including previously unseen footage, to generate renewed interest and boost ticket sales. [D101] \textsc{Outcome: \#2}
    \item The new version of the film did not include any re-edited scenes within the main feature but instead featured additional scenes after the end credits, including a tribute. [D14] \textsc{Outcome: \#3, Nugget: 3}
    \item This approach aimed to entice both existing and new audiences to revisit the theaters and contribute to the film's box office success. \textsc{Outcome: \#6}

\end{enumerate} 
\vspace*{-\baselineskip}
& {} &
\vspace*{-0.5\baselineskip}
\begin{enumerate}[leftmargin=*]
    \setcounter{enumi}{8}
    
    \item Similarly, Avatar also made a comeback to reclaim its position as the highest-grossing film. \textsc{Outcome: \#4}
    \item The film was re-released in China, where it achieved a single-day box office revenue of \$8 million. [D2] \textsc{Outcome: \#3, Nugget: 5}
    \item The re-release in China was particularly successful, as it generated significant revenue and propelled Avatar back to the top of the global box office charts. [D1] \textsc{Outcome: \#3, Nugget: 5}
    \item The film's re-release in China was attributed to the impact of the COVID-19 pandemic, which led to the delay of many new releases and created an opportunity for Avatar to regain its popularity. [D2] \textsc{Outcome: \#3, Nugget: 5}
    \item In conclusion, Avengers: Endgame and Avatar both held the title of the highest-grossing film globally at different points in time. \textsc{Outcome: \#4}
    \item Marvel Studios strategically re-released Avengers: Endgame with additional content to attract audiences and boost ticket sales. \textsc{Outcome: \#6}
    \item Avatar capitalized on the re-release trend in China, taking advantage of the pandemic-induced delay of new releases. \textsc{Outcome: \#6}
    \item These manipulations aimed to increase the films' box office revenue and solidify their positions as record-breaking blockbusters. \textsc{Outcome: \#4}

\end{enumerate}
\vspace*{-\baselineskip} \\
\hline

\end{tabular}

\vspace{-1em}    
\caption{Example report evaluation result. } \label{fig:example-eval-result}
\end{figure*}

Recall and precision each require a numerator and a denominator.
The recall denominator is the number of distinct assessor-identified nuggets;
its numerator is the number of correctly reported nuggets
(those supported by one or more of the necessary supporting citations in the report).
So recall tells us how many of the concepts central to the report were actually reported on.
Precision must account for  phenomena below the nugget level,
so we calculate it over \sourceelement s
(which again we assume to be sentences).
The denominator is the number of report sentences,
minus any sentence that does not require a citation
or that properly cites information not part of any nugget. %
The numerator is the number of sentences deemed to bear accurate citations,
plus any sentences specified by the evaluation that correctly bear no citation.

Figure~\ref{fig:scoring} describes a typical approach to sentence evaluation.
The rules embedded in the flowchart are not hard-and-fast,
but are likely adequate for many evaluations under this framework.
In the flowchart, ``+'' means the sentence is correct and should be rewarded;
``--'' means that it is incorrect and should be penalized; 
and ``0'' means that the sentence is not included in the score.
The flowchart shows how each sentence of the report can be scored.
We propose these principles to guide sentence scoring:
\begin{itemize}[noitemsep,topsep=0pt]
    \item Sentences with citations whose \targetelement\ does not support them should be penalized
    (Outcome\#1 in Figure~\ref{fig:scoring}).
    \item Properly cited and attested sentences that are not relevant to the report should be ignored (Outcome \#2).
    \item A sentence that cites a \targetelement\ supporting a nugget that the sentence fulfills should be rewarded (Outcome \#3).
    \item Sentences that neither have nor require citations should not affect the score (Outcome \#4).%
    \item Sentences that should contain a citation but do not should be penalized the first time their claim occurs (Outcomes \#5, \#6).
    \item Sentences that claim the absence of a fact should be rewarded or penalized
    depending on whether the absence is explicitly stated as a nugget (Outcomes \#7, \#8). %
    For this, a nugget can be created for information that the \reportrequest\ explicitly asks for but is not attested in the collection.
\end{itemize}

\noindent Most sentences will bear either zero or one citation.
A sentence can bear multiple citations,
either because the same information is multiply attested in the collection,
or because it is complex.
Sentences that cite multiple \targetelement s supporting the same nugget
are treated as a single citation.
Alternatively, the evaluation may macroaverage citation scores
if all sentences are to be given equal weight,
or microaverage them if the focus is on citation accuracy.
Support by multiple report sentences counts only once per nugget.

To automatically score a report, each decision diamond in Figure~\ref{fig:scoring} must be automatable.
Some are trivial, such as ``Has citation;''  others are less so. 
We believe current technology could do a reasonable job with most of the tasks.
For instance, entailment models can likely determine if a document supports a report sentence.
Note that originality is not a component of this evaluation;
preventing plagiarism, while important,
is a specialized area with its own metrics and evaluations~\cite{potthast-plagiarism,belyy-plagiarism,foltynek-plagiarism,jambi-plagiarism}.

\section{Example Assessment}\label{sec:example}

Figure~\ref{fig:example-eval-info} shows an example of the two items
required to do manual or automatic assessment.
The \reportrequest\ identifies the desired report content.
The nugget questions and answers
show how each answer is linked to the documents that attest to that answer.
The Gold Standard Report
that is shown is optional, but a useful intermediate step for the assessor between
source document search and nugget question creation.

Figure~\ref{fig:example-eval-result} is a report generated in response to the example in Figure~\ref{fig:example-eval-info},
broken into report segments to illustrate manual evaluation.
Each \textsc{Outcome: \#}
indicates how the sentence would be categorized using the flowchart in Figure~\ref{fig:scoring}.
For \textsc{Outcome: \#3}, the nugget answer in the sentence is also recorded.
No sentence received a negative assessment because there were no outcomes of \textsc{\#1} or \textsc{\#7}.
Therefore, precision is $5/(16-11) = 1.0$.
One nugget was repeated in Lines 10, 11, and 12, so recall is $3/5 = 0.6$.
For both Lines 2 and 6,
the assessor would have needed to refer to the original source document to assess the statement,
since the information in the sentence had not been captured in a required nugget.
Assessing such sentences will likely be the most time-consuming part of manual assessment. 
\section{Conclusions}\label{sec:conclusion}

LLMs have enabled remarkable new ways to satisfy information needs.
Rather than simply providing ``10 blue links'' or an extracted answer snippet,
LLMs have the potential to peer into documents to identify information salient to a topic
and compile it into highly coherent, long-form text responses.
We envision these generated reports will be a central way
that some users will satisfy complex, nuanced, or multifaceted information needs.

Because we believe that current evaluation methodologies for these report-generation systems are insufficient
to maintain quality and guard against known defects,
we felt the need for a report evaluation framework
based on core principles ---
responsiveness to the information need,
grounding and verifiability in documents,
completeness, and reusability ---
while deliberately omitting aspects of report generation that current systems do not seem to struggle with
(e.g., coherence, structure, etc.).
Our new perspective on report generation evaluation is IR-centric,
pulling together tried-and-true notions of relevance, recall, and user modeling.
We have also demonstrated an instantiation of our framework
that could be applied either manually or with automatic systems.

Evaluation methodologies inform progress and direct attention.
We hope our proposed generated report evaluation framework will spur progress
in the development of next-generation information access systems
that can provide responsive, complete, and verifiable information
on complex, nuanced, and multifaceted topics.

\section*{Whose Perspective}

This paper represents the perspectives of a group of industry and academic researchers at a variety of career stages.
We thank John Conroy, Hoa Dang, Mohit Iyyer, Jimmy Lin, and Graham Neubig for their discussions with us.

\bibliographystyle{ACM-Reference-Format}
\bibliography{bib}

\appendix

\end{document}